\documentclass[runningheads]{llncs}
\usepackage{graphicx}
\usepackage{epstopdf}
\usepackage{color}
\usepackage{cite}
\usepackage{amsfonts,amssymb,amsmath}
\usepackage{multirow}
\usepackage{epsf}
\usepackage{longtable}
\usepackage{booktabs}
\usepackage[table]{xcolor}
\usepackage{ragged2e}
\usepackage{subfigure}
\usepackage{marvosym}
\usepackage{rotating}
\usepackage[utf8]{inputenc}  
\DeclareUnicodeCharacter{FF0C}{,}  

\begin{document}
\title{FusionFM: Fusing Eye-specific Foundational Models for Optimized Ophthalmic Diagnosis}
\author{Ke Zou\inst{1,2}, Jocelyn Hui Lin Goh\inst{1,2,3}, Yukun Zhou\inst{5}, Tian Lin\inst{6}, Samantha Min Er Yew\inst{1,2}, Sahana Srinivasan\inst{1,2}, Meng Wang\inst{1,2}, Rui Santos\inst{7,8}, Gabor M. Somfai\inst{7,8}, Huazhu Fu\inst{9}, Haoyu Chen\inst{6}, Pearse A. Keane\inst{10,11}, Ching-Yu Cheng\inst{1,2,3,4},  Yih Chung Tham\inst{1,2,3,4}\textsuperscript{\Letter}}
\authorrunning{K. Zou et al.}
\institute{Department of Ophthalmology, Yong Loo Lin School of Medicine, National University of Singapore, Singapore \and Centre for Innovation and Precision Eye Health, Yong Loo Lin School of Medicine, National University of Singapore, Singapore \and
Singapore Eye Research Institute, Singapore National Eye Centre, Singapore, Singapore \and Ophthalmology and Visual Science Academic Clinical Program, Duke-NUS Medical School, Singapore, Singapore \and Centre for Medical Image Computing, University College London, London, United Kingdom \and Joint Shantou International Eye Center, Shantou University and the Chinese University of Hong Kong, Shantou, China \and Department of Ophthalmology, Stadtspital Zürich, Zurich, Switzerland \and Spross Research Institute, Zurich, Switzerland \and Institute of High Performance Computing, A*STAR, Singapore \and NIHR Biomedical Research Centre, Moorfields Eye Hospital NHS Foundation Trust, London, United Kingdom \and  Institute of Ophthalmology, University College London, London, United Kingdom\\
\email{thamyc@nus.edu.sg}}
\maketitle              
\begin{abstract}
Foundation models (FMs) have shown great promise in medical image analysis by improving generalization across diverse downstream tasks. In ophthalmology, several FMs have recently emerged, but there is still no clear answer to  fundamental questions: \textit{Which FM performs the best? Are they equally good across different tasks? What if we combine all FMs together?} To our knowledge, this is the first study to systematically evaluate both single and fused ophthalmic FMs. To address these questions, we propose FusionFM, a comprehensive evaluation suite, along with two fusion approaches to integrate different ophthalmic FMs. Our framework covers both ophthalmic disease detection (glaucoma, diabetic retinopathy, and age-related macular degeneration) and systemic disease prediction (diabetes and hypertension) based on retinal imaging. We benchmarked four state-of-the-art FMs (RETFound, VisionFM, RetiZero, and DINORET) using standardized datasets from multiple countries and evaluated their performance using AUC and F1 metrics. Our results show that DINORET and RetiZero achieve superior performance in both ophthalmic and systemic disease tasks, with RetiZero exhibiting stronger generalization on external datasets. Regarding fusion strategies, the Gating-based approach provides modest improvements in predicting glaucoma, AMD, and hypertension. Despite these advances, predicting systemic diseases—especially hypertension in external cohorts—remains challenging. These findings provide an evidence-based evaluation of ophthalmic FMs, highlight the benefits of model fusion, and point to strategies for enhancing their clinical applicability.
\end{abstract}

\section{Introduction}
Foundation models (FMs) leverage large-scale datasets and pretraining strategies to build versatile models capable of generalizing across a wide range of downstream tasks. In recent years, medical foundation models have achieved significant success across various imaging domains, including radiology~\cite{ma2025fully,bluethgen2024vision}, pathology~\cite{xu2024whole,wang2024pathology}, oncology~\cite{xiang2025vision}, and dermatology~\cite{dermatology2025}, demonstrating improved accuracy and generalization on diverse clinical tasks.

Compared to these fields, ophthalmology presents unique opportunities for FM development. Ocular images, such as fundus photographs and optical coherence tomography (OCT) scans, capture rich biomarkers not only for eye diseases but also for systemic conditions, making them a valuable modality for developing generalizable medical foundation models. With the growing availability of ophthalmic imaging data and advancements in model architectures, several ophthalmic FMs~\cite{zhou2023retfound,qiu2023visionfm,shi2024eyefound,retizero2024,zoellin2024block,da2025integrated,liu2024octcube,peng2025enhancing,wong2025eyefm} have recently emerged, highlighting the potential of foundation models to support a broad range of clinical applications in eye care and beyond. In the following, we provide an overview of four representative FMs in ophthalmology, each characterized by distinct model architectures and pretraining strategies. First, the pioneering retinal foundation model RETFound~\cite{zhou2023retfound} learns generalizable representations from large-scale unlabeled retinal images, providing a strong foundation for adapting to various ophthalmic applications. Then, VisionFM~\cite{qiu2023visionfm} leverages pretraining across diverse ophthalmic modalities and imaging devices to establish a strong foundation for disease screening and diagnosis, prognosis prediction, disease phenotyping and subtyping tasks. Furthermore, RetiZero~\cite{retizero2024} trains a knowledge-enriched vision-language foundation model using public datasets, ophthalmology literature, and online resources, enabling zero-shot disease recognition, image-to-image retrieval, AI-assisted clinical diagnosis, few-shot fine-tuning, and both in-domain and cross-domain disease identification. More recently, DINORET~\cite{zoellin2024block} employs DINOv2~\cite{oquab2023dinov2} vision transformers for self-supervised learning on retinal image classification tasks and addresses the issue of catastrophic forgetting during foundation model fine-tuning. While these models have demonstrated strong potential across a range of clinical applications, systematic benchmarking and comparative analysis remain scarce. Yew et al.\cite{yew2025traditional} evaluated RETFound against traditional CNN-based approaches, but direct, head-to-head comparisons among modern foundation models are still lacking. Zhou et al.\cite{zhou2025revealing} investigated the composition of pretraining datasets across various FMs, highlighting substantial differences in data sources and scales. However, a comprehensive and systematic comparison of these models under unified evaluation settings is still lacking. This limitation is further compounded by the fact that existing FMs are often pretrained on heterogeneous datasets with varying curation pipelines and annotation quality. Moreover, the absence of a standardized benchmarking framework and consistent evaluation protocols makes it difficult to fairly assess and compare model performance across ophthalmic and systemic disease tasks. In addition, it remains unclear whether combining multiple FMs can improve performance, and if so, which strategies are most effective for fusion in the context of basic ophthalmic models.

To address these limitations, we propose FusionFM, a comprehensive benchmarking suite covering key tasks in both ophthalmic disease detection and systemic disease prediction from color fundus photographs. FusionFM further explores different fusion strategies and introduces a novel approach to integrate multiple ophthalmic FMs using two methods, Gating-based and Router-based, enabling the first systematic investigation of whether combining models can enhance performance. Specifically, we curated and organized datasets collected from both public and private sources for core ophthalmic disease detection tasks, including glaucoma, diabetic retinopathy (DR), and age-related macular degeneration (AMD), which have not been previously used for pretraining existing FMs. In addition, we constructed new datasets for systemic disease prediction tasks, such as diabetes and hypertension, where relevant biomarkers can be inferred from retinal images. To ensure fair comparisons across different foundation models, FusionFM employs widely used evaluation metrics to assess performance across a broad range of disease categories and tasks. Our experiments demonstrate that DINORET and RetiZero achieve strong performance in both ophthalmic and systemic disease tasks, with RetiZero showing superior generalization on external datasets. The Gating-based fusion strategy provides small but consistent improvements in predicting glaucoma, AMD, and hypertension. Nevertheless, accurately predicting systemic diseases, particularly hypertension in external cohorts, remains a challenging problem that warrants further investigation.

\section{Design of FusionFM}
\begin{figure*}[!t]
\centering
\includegraphics[width=1\linewidth]{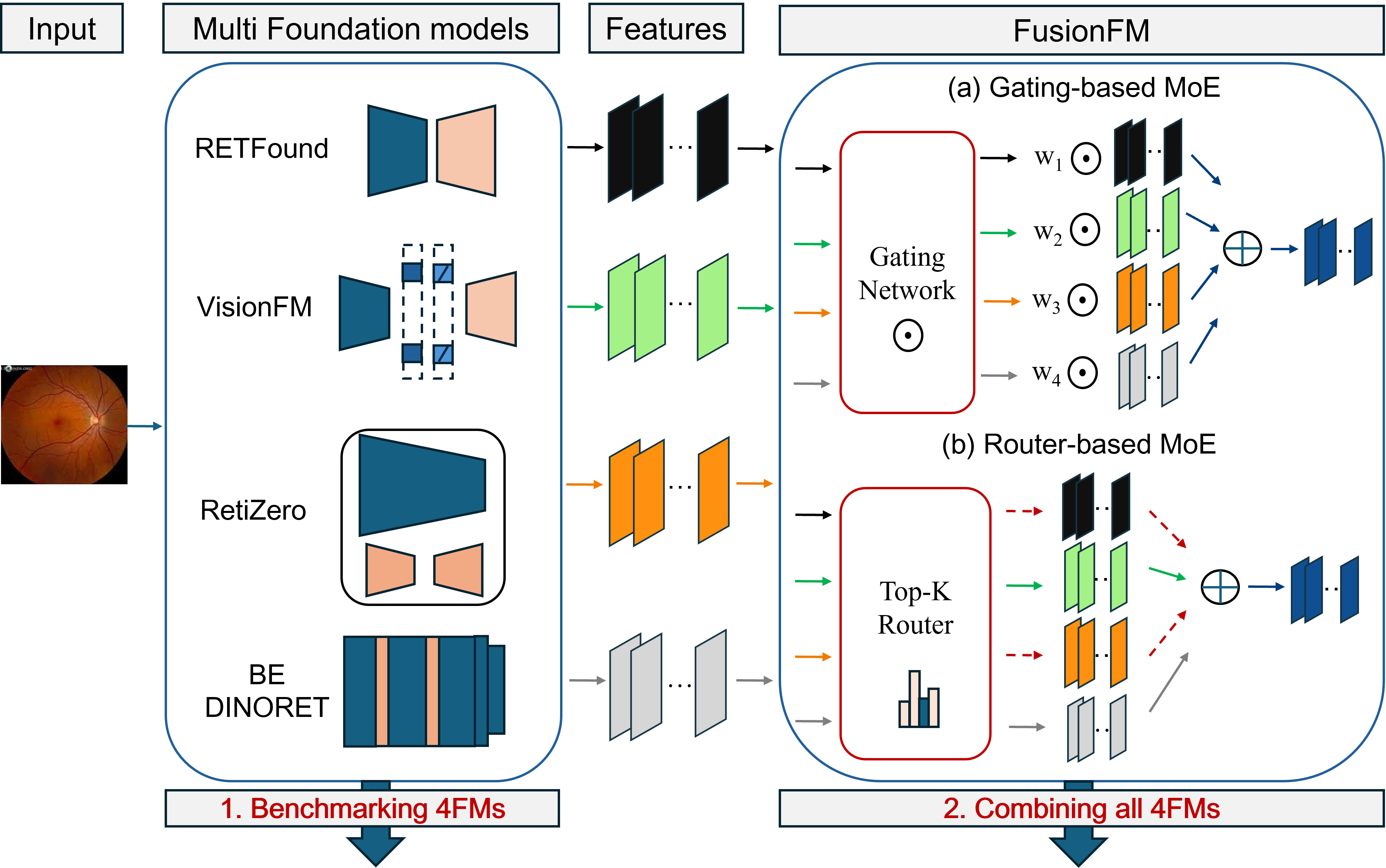}
\includegraphics[width=1\linewidth]{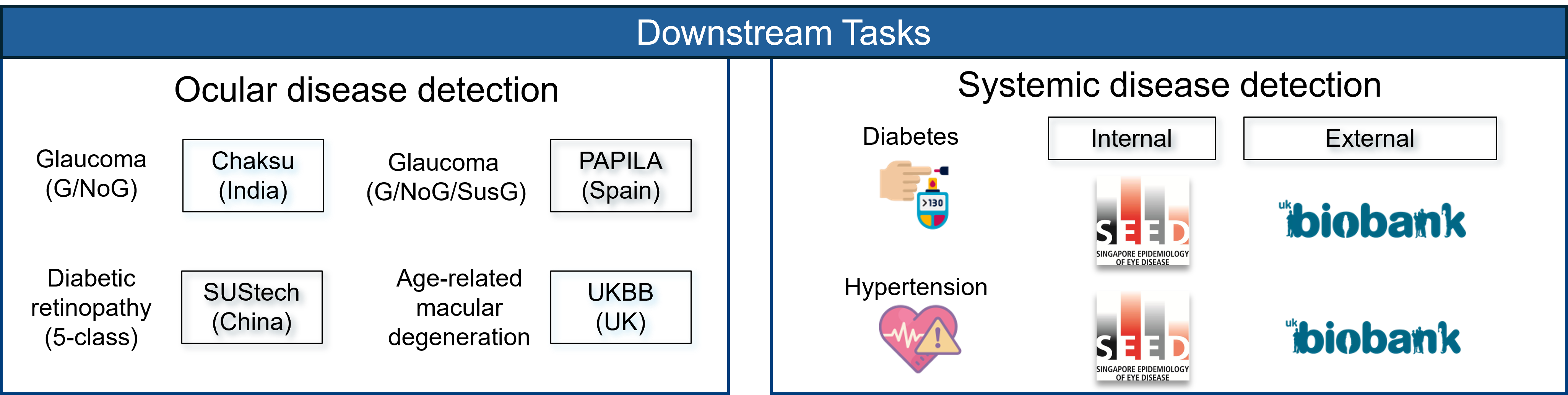}
\caption{Overview framework of EyeFusion.}
\label{F_1}
\end{figure*}

As shown in the Fig.~\ref{F_1} illustrates the overall framework of the proposed FusionFM. The core design follows a frozen backbone strategy, where only the classification head is trained for downstream tasks. The benchmark covers both ophthalmic disease detection and systemic disease prediction, with model performance evaluated through comprehensive metrics.

\subsection{Foundation models for Ophthalmology} 
Four representative ophthalmic foundation models were employed for comparison on this benchmark, each pre-trained with different corpora, image modalities, and strategies. RETFound adopts a masked autoencoder (MAE)-based self-supervised learning framework using large-scale unlabeled retinal fundus and OCT images. Specifically, it is pre-trained on 1.64 million images across two modalities from the EyePACS and MEH-MIDAS datasets. VisionFM is built upon an iBOT-based architecture, employing self-supervised learning with image-token reconstruction and self-distillation. It is pre-trained on 3.4 million images spanning eight ophthalmic modalities, collected from various public and private datasets. The model adopts separate encoders for each modality and incorporates task-specific decoders to support a wide range of clinical applications. RetiZero employs a multimodal architecture that combines a masked autoencoder (MAE) with Low-Rank Adaptation (LoRA) for the image encoder and a CLIP-based text encoder enhanced with uncertainty estimation. Its pretraining corpus consists of approximately 340K paired image-text samples from 30 public ophthalmology datasets, covering two imaging modalities. DINORET leverages a block expansion strategy to adapt pre-trained DINOv2 vision transformers to the ophthalmic domain. It is trained using more than 202K color fundus images sourced from four public datasets. The model inserts identity-initialized duplicate transformer blocks into DINOv2 and fine-tunes only these newly added blocks using contrastive self-supervised learning, thereby retaining general visual representations while adapting to ophthalmic-specific characteristics. For consistency, all four foundation models adopt a data augmentation pipeline similar to that of RETFound.

\subsection{Settings \& Data Sources}
\textbf{1) Problem formulation.} To ensure a fair comparison across different foundation models, we adopt a unified experimental setting. For each model, we load its pre-trained backbone weights \( f(w_i) \), where \( w_i \) denotes the frozen parameters of the \( i \)-th foundation model. We then train only a lightweight task-specific classification head, denoted as \( f^c_i(\cdot) \), as illustrated in Figure~\ref{F_1}. Given an input fundus image \( x \), the prediction for a target task (either ophthalmic disease detection or systemic disease prediction) is defined as:
\begin{equation}
    \hat{y} = f^c_i \big( f(w_i)(x) \big)
\end{equation}
Here, \( \hat{y} \in \mathbb{R}^K \) denotes the predicted logits or probabilities for \( K \) classes in the given task. Only the classification head \( f^c_i \) is optimized during training, while the backbone \( f(w_i) \) remains frozen. This unified setup allows us to evaluate the representational quality of different foundation models under identical conditions on two types of tasks. In the following sections, we introduce the detailed task settings and the corresponding datasets used for each task. These include both publicly available and private datasets, covering diverse populations and imaging devices to ensure robustness and generalizability in evaluation.\\
\textbf{2) Task 1: Ocular diseases.}  Glaucoma, DR, and AMD are among the most common and clinically significant ophthalmic diseases worldwide. These conditions are major causes of visual impairment and blindness, posing a substantial global health and socioeconomic burden. Given their high prevalence, well-established diagnostic criteria, and the availability of annotated datasets, these diseases serve as essential benchmark tasks for evaluating the effectiveness and generalizability of ophthalmic foundation models. Therefore, FusionFM includes glaucoma, DR, and AMD as representative tasks for standardized assessment of model performance in this domain.\\
For glaucoma detection, we utilize the publicly available Chaksu~\cite{kumar2023chakṣu} and PAPILA~\cite{kovalyk2022papila} datasets. The Chaksu dataset was developed as an Indian ethnicity fundus image database for glaucoma prescreening. It contains a total of 1,345 color fundus images acquired using three different brands of commercially available fundus cameras. For each image, experts provided binary glaucomatous or non-glaucomatous labels, determined through majority voting to ensure a single consensus label per image. The dataset includes 188 glaucomatous and 1,157 non-glaucomatous images. The PAPILA dataset was curated at the Department of Ophthalmology, Hospital General Universitario Reina Sofía (HGURS), Murcia, Spain, with data collected between 2018 and 2020. It includes records from 244 patients, comprising a total of 488 fundus images in JPEG format, corresponding to both the right and left eyes. All images are categorized into three classes: healthy (333 images), glaucoma (68 images), and glaucoma suspect (87 images). For the DR task, we adopt the SUSTech-SYSU dataset~\cite{lin2020sustech}, which contains 1,219 fundus images collected from DR patients and healthy controls at Gaoyao People’s Hospital and Zhongshan Ophthalmic Center, Sun Yat-sen University. The dataset includes 631 normal images and 588 DR images annotated across five severity levels: mild non-proliferative DR (24 images), moderate non-proliferative DR (365 images), severe non-proliferative DR (73 images), proliferative DR (58 images), and DR with laser spots or scars (68 images). For the AMD task, we use the accessible UK Biobank (UKBB) dataset, which is categorized into two classes: AMD and others, with approximately 680 and 9.3K cases respectively. For all aboved tasks, the corresponding datasets were split into training, validation, and test sets with a ratio of 7:1:2. \\
\textbf{3) Task 2: Systematic diseases.} Diabetes and hypertension are among the most common systemic diseases globally, contributing substantially to morbidity and mortality. Both conditions exhibit well-characterized ocular manifestations that can be captured through retinal imaging, a field often referred to as oculomics. Early detection of these diseases via retinal biomarkers enables timely intervention and improved patient outcomes. The Singapore Eye Epidemiology (SEED) cohort is a multi-ethnic longitudinal population study designed to investigate the morbidity, prevalence, risk factors, and novel biomarkers of age-related eye diseases among Malay, Indian, and Chinese populations in Singapore. SEED provides a valuable resource for evaluating ethnic differences in the incidence and progression of ocular and systemic diseases. Similarly, the UKBB is a large-scale multi-ethnic longitudinal population study that includes extensive phenotypic and genetic labels for a wide range of systemic diseases. In this study, we selected diabetes and hypertension from SEED for internal validation, and the same disease labels from UKBB for external validation. The UKBB dataset includes approximately 5K diabetes cases and 12K controls, and approximately 7K hypertension cases and 11K controls.

\subsection{Fusion Strategies for Multiple Ophthalmic Foundation Models}
1) \textbf{Multi-Foundation Model Feature Extraction:} To harness the complementary strengths of multiple pretrained ophthalmic foundation models, we first loaded each model and extracted intermediate features from their respective backbones. These feature representations capture rich semantic cues derived from diverse training objectives and data distributions. The extracted features were then aligned in both channel and spatial dimensions and fused to construct a unified representation. This fusion enables downstream models to benefit from the ensemble knowledge of diverse inductive biases embedded across different foundation models.\\
2) \textbf{Gating-based MoE strategy:} To effectively integrate heterogeneous features from multiple foundation models, we first developed a gating-based MoE strategy tailored for ophthalmic FMs fusion. This approach introduces a lightweight gating network  that adaptively learns to balance and combine intermediate features from each model, generating a fused representation conditioned on the input image. By dynamically assigning importance to each expert, the gating mechanism enables the proposed method to leverage complementary knowledge information across FMs, thereby enhancing generalization across diverse ophthalmic tasks.\\
3) \textbf{Top-K Router-based MoE strategy:} Furthermore, we explored a Top-K router-based MoE strategy, which provides a sparse and efficient mechanism for expert selection. Unlike gating-based approaches that aggregate outputs from all experts, the Top-K routing method selects only the K most relevant experts based on the input, significantly reducing computational overhead while maintaining predictive performance. This selective activation encourages specialization among experts and facilitates effective utilization of diverse feature representations across foundation models.

\subsection{Evaluation protocol \& training details}
In addition to the commonly used area under the receiver operating characteristic curve (AUC), which has been adopted by RETFound, VisionFM, and RetiZero, we also included accuracy (ACC) as the evaluation metric, following the practice of DINORET. This allows for a more comprehensive assessment of the performance of all four foundation models on both ophthalmic disease detection and systemic disease prediction tasks. \\
To ensure a fair comparison across all foundation models, we adopted a unified training loss function for all experiments. Specifically, all foundation models were trained using the label smoothing Cross entropy loss, which mitigates overconfidence in predictions and promotes better generalization. The loss function is defined as:
\begin{equation}
\mathcal{L} = - (1 - \epsilon) \log(p_{y}) - \frac{\epsilon}{K} \sum_{i=1}^{K} \log(p_{i})
\label{E_1}
\end{equation}
where \( p_{y} \) is the predicted probability of the ground-truth class, \( K \) is the total number of classes, and \( \epsilon \) is the smoothing parameter. In all experiments, we set the label smoothing parameter \(\epsilon\) to 0.1 following RETFound~\cite{zhou2023retfound}, which helps prevent overfitting and promotes better generalization.

\section{Experiments \label{S_5}}
\subsection{Implementation Details}
All experiments were conducted on the PyTorch platform using an NVIDIA A6000 GPU with 48 GB of memory. We adopted a layer-wise learning rate decay strategy when constructing the optimizer, grouping model parameters with different decay rates based on their depth within the network. All models were optimized using the AdamW optimizer with an initial learning rate of 0.0001. The maximum number of training iterations was set to 100, and the batch size was fixed at 16. During training, different random seeds were used, and the checkpoint with the best AUC on the validation set was saved for all baselines. All experiments employed a bootstrap method with 1,000 sampling iterations to obtain 95\% confidence intervals (CIs). All images are resized to $224\times224$ during fine-tuning. 

\subsection{Experimental Results}
\begin{figure*}[!t]
\centering
\includegraphics[width=1\linewidth]{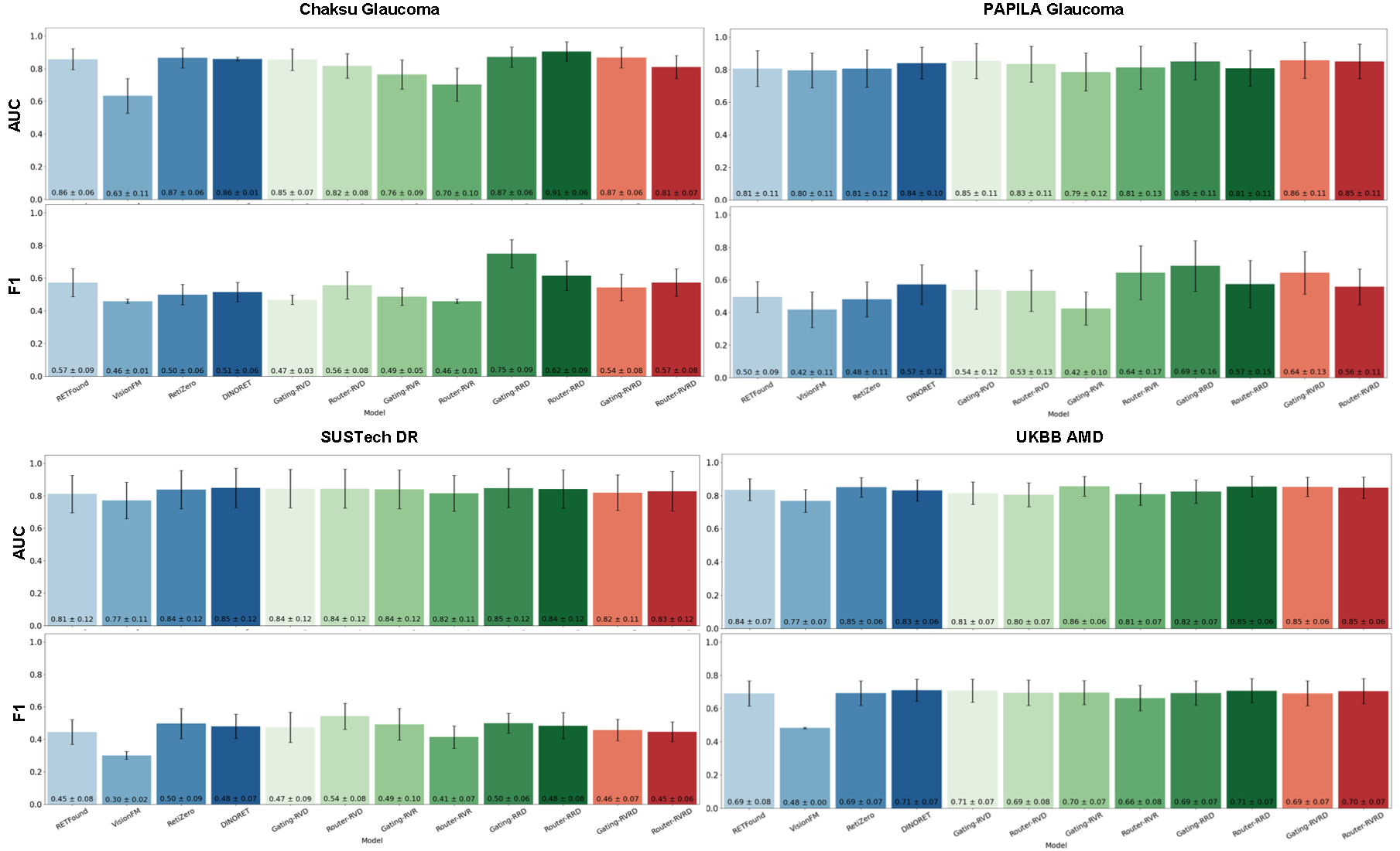}
\caption{Performance (AUC and F1) of four foundational models and various fusion strategies within the FusionFM framework for ocular disease detection, evaluated across multiple datasets.}
\label{F_2}
\end{figure*}

\begin{figure*}[!t]
\centering
\includegraphics[width=1\linewidth]{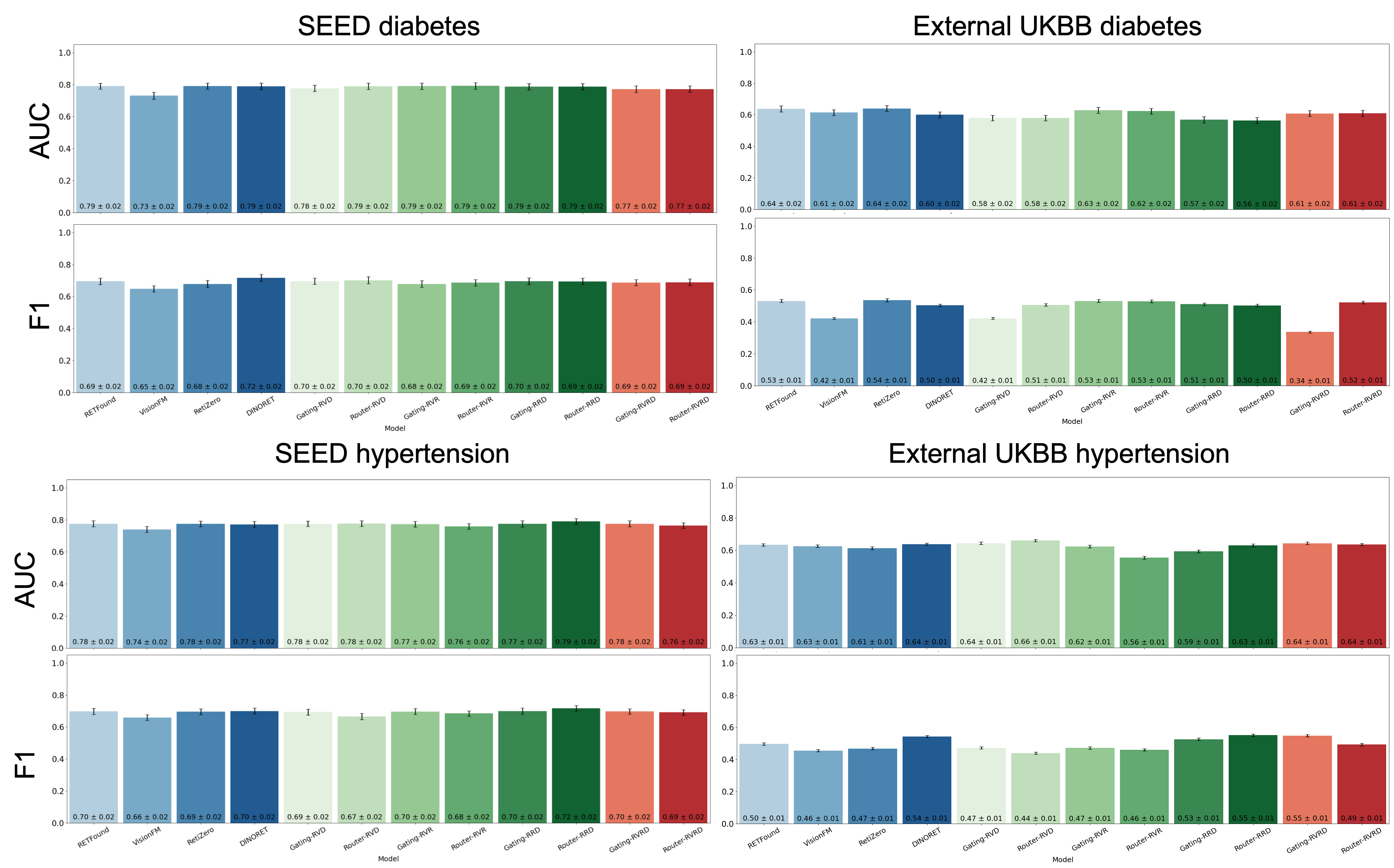}
\caption{Performance (AUC and F1) of four eye-specific foundation models and various FusionFM strategies for systemic disease detection from ocular images, evaluated on the SEED and UKBB datasets.}
\label{F_3}
\end{figure*}

Figures~\ref{F_2} and~\ref{F_3} summarize the performance of the four foundation models on ophthalmic and systemic disease detection tasks across different datasets, using ACC and AUC as evaluation metrics. The experimental results are further elaborated and analyzed in the following sections.\\
\textbf{Results of ocular diseases.} As shown in Fig.~\ref{F_2}, we present the performance of different foundation models on three ophthalmic diseases across four datasets from different countries. In terms of AUC with 95\% CIs, for glaucoma detection on the Chaksu and PAPILA datasets, DINORET and RetiZero achieved comparable performance, both outperforming VisionFM. On the PAPILA dataset, DINORET also outperformed the other FMs. For DR detection (SUSTech dataset) and AMD detection (UKBB dataset), a similar pattern was observed: RETFound and RetiZero again achieved comparable performance and outperformed VisionFM. In terms of F1 with a 95\% CI, the four baseline models, when applied to the Chaksu and PAPILA datasets, achieved superior performance with RETFound or DINORET in glaucoma detection. For DR detection on the SUSTech dataset, RetiZero performed better, while DINORET achieved the highest performance on the AMD task using the UKBB dataset. For the fusion strategy, Gating-RRD and Router-RRD achieved the best performance on the glaucoma detection task (Chaksu dataset), outperforming the four FMs. The best fusion strategy exceeded the optimal FM by 4\% in AUC and 18\% in F1. Similarly, on the PAPILA dataset, Gating-RRD achieved the highest performance, surpassing the four FMs with a 12\% higher F1 score. For DR classification, the fusion strategy performed comparably to the best FM. In the AMD task, when considering both AUC and F1 metrics, Gating-RVR slightly outperformed the best FMs.

Overall, these results indicate that RETFound, RetiZero, and DINORET tend to outperform VisionFM in ophthalmic disease detection. One possible reason is that VisionFM is based on a ViT architecture, whereas RETFound and RetiZero both share the same ViT backbone architecture. Moreover, DINORET may benefit from the diversity of its pre-training datasets, which include various ophthalmic datasets, as well as from the powerful DINOv2 backbone. For the fusion of different FMs, in glaucoma and AMD detection tasks, fusing multiple FMs tends to yield better performance. This superior performance is likely due to the complementary strengths of individual models in capturing multi-scale lesion features, adapting to varying image distributions, and reducing prediction uncertainty through gating- or router-based expert selection.\\
\textbf{Results of systematic diseases.} As shown in Fig.~\ref{F_3}, we compared the performance of different baseline models on two systemic disease prediction tasks using datasets from two countries. Based on AUC and F1 with a 95\% confidence interval, diabetes prediction on the SEED internal dataset showed that DINORET, RetiZero, and RETFound performed equally well and all outperformed VisionFM. For the fusion strategies, performance was comparable to that of the best FM in both AUC and F1. On the external validation dataset (UKBB, Column 2), the FMs showed similar AUC values, with DINORET having a slight advantage in F1. Among the fusion strategies, Router-RRD and Gating-RVRD achieved slightly higher AUC, while Gating-RVRD also demonstrated a marginal improvement in F1.
Overall, DINORET and RetiZero performed better in systemic disease detection. The fusion strategy results further indicate that fusion is beneficial in the hypertension experiment. Moreover, compared with the internal dataset, performance on the external dataset was lower, highlighting the challenges of hypertension prediction in an external cohort.

Overall, these results indicate that DINORET, RetiZero, and RETFound generally outperformed VisionFM in systemic disease prediction tasks, with RetiZero demonstrating stronger generalization in diabetes prediction. One possible reason is that RetiZero benefited from a richer and more diverse pre-training dataset, including public datasets, books, and online resources, thereby enhancing its generalization ability. In addition, its fine-tuning strategy combines RETFound with LoRA. On the other hand, DINORET demonstrated a slight edge in both diabetes and hypertension prediction on the internal validation set, which might be attributed to the powerful DINOv2 backbone and its pre-training on ophthalmic datasets.
Regarding the fusion of different FMs, the experimental results show that for hypertension prediction, the Gating-based strategy that fuses all models can achieve a slight advantage over the optimal FM. This improvement may benefit from the complementary strengths of individual FMs in capturing multi-scale retinal features. Despite this, all models struggled to predict hypertension on external datasets, highlighting an area worthy of further study. In conclusion, our results indicate that DINORET and RetiZero perform comparably to RETFound on internal datasets for ophthalmic and systemic disease prediction tasks, while providing better generalization on external datasets. For different fusion strategies, the current experiments suggest that a Gating-based fusion strategy is recommended for glaucoma detection, AMD detection, and prediction of systemic hypertensive diseases.

\section{Conclusions}
In this study, we evaluated representative foundational models (FMs) in ophthalmology and explored various fusion strategies. To our knowledge, this is the first study systematically investigating both single and fused ophthalmic FMs, with the aim of optimizing their detection capabilities across multiple tasks. We further proposed two Mixture-of-Experts approaches, namely the Gating-based and Router-based methods.
Our experimental results demonstrate that DINORET and RetiZero achieve superior performance in both ophthalmic disease detection and systemic disease prediction. Specifically, DINORET shows strong potential in ophthalmic and systemic tasks, while RetiZero exhibits better generalization on external datasets, emphasizing the importance of diverse pre-training data and effective fine-tuning strategies. Regarding fusion, the results suggest that a Gating-based strategy is advantageous for glaucoma detection, AMD detection, and hypertension prediction. Despite these advances, predicting systemic diseases， particularly hypertension in external cohorts—remains a challenging problem. In future work, we plan to further analyze model architectures and efficiency to enhance transparency and interpretability of each FM. Additionally, we advocate for broader clinical deployment of these models to maximize their real-world impact.
 
%
%
%
\bibliography{paper_ref}
\bibliographystyle{splncs04}

\end{document}